\documentclass{article}

\usepackage{graphicx}
\usepackage{hyperref}
\usepackage{wrapfig}
\usepackage{textcomp}
\usepackage{gensymb}
\usepackage{amsmath}
\usepackage{amssymb}
\usepackage[ruled,vlined,linesnumbered,noresetcount]{algorithm2e}
\usepackage{algpseudocode}
\usepackage[utf8]{inputenc}
\usepackage[xindy]{glossaries}
\usepackage[toc,page]{appendix}
\usepackage{multirow}
\usepackage{booktabs}
\usepackage{lscape}
\usepackage{tabulary}
\usepackage{subcaption}
\usepackage[dvipsnames]{xcolor}
\usepackage{color,soul} 
\usepackage{float} 
\usepackage{microtype} 
\usepackage{multicol}
\usepackage[preprint]{corl_2020} 

\usepackage{lipsum}
\usepackage{libertine}

\newenvironment{Figure}
  {\par\medskip\noindent\minipage{\linewidth}}
  {\endminipage\par\medskip}
  
\usepackage[symbol]{footmisc}

\title{Iterative Semi-parametric Dynamics Model Learning For Autonomous Racing}

\author{
    Ignat Georgiev\footnote[2]{} \footnote[3]{} \space,  
    Christoforos Chatzikomis\footnote[2]{} \space,  
    Timo Völkl\footnote[2]{} \space \footnotemark \space,
    Joshua Smith\footnote[3]{} \space,
    Michael Mistry\footnote[3]{}\\ 

    \footnote[2]{} \space Arrival\\
    \footnote[3]{} \space School of Informatics, The University of Edinburgh\\
}

\begin{document}
\maketitle


\begin{abstract}
Accurately modeling robot dynamics is crucial to safe and efficient motion control. In this paper, we develop and apply an iterative learning semi-parametric model, with a neural network, to the task of autonomous racing with a Model Predictive  Controller (MPC). We present a novel non-linear semi-parametric dynamics model where we represent the known dynamics with a parametric model, and a neural network captures the unknown dynamics. We show that our model can learn more accurately than a purely parametric model and generalize better than a purely non-parametric model, making it ideal for real-world applications where collecting data from the full state space is not feasible. We present a system where the model is bootstrapped on pre-recorded data and then updated iteratively at run time. Then we apply our iterative learning approach to the simulated problem of autonomous racing and show that it can safely adapt to modified dynamics online and even achieve better performance than models trained on data from manual driving.
\end{abstract}



\section{Introduction}
\label{sec:intro}

\footnotetext{Author's affiliation has changed since this research was done. New affiliation with Munich Electrification}

Robots are one of the most fascinating human-made machines. They raise hopes of exploring extraterrestrial planets, replacing human workers in dangerous environments, assisting us with laborious tasks, and even saving human lives \citep{siegwart2011introduction}. One of the first instances of potential for massive adaption of this technology has been autonomous cars, which are promising to alleviate traffic congestion, reduce road incidents, and lower pollution \cite{fagnant2015preparing}. A challenging aspect of driverless vehicle development is controlling them accurately, safely, and efficiently. One popular approach to solve this problem is using model-based control, where the system dynamics are modeled and then used by some controller to achieve the desired task. In recent years, the most promising direction for these systems has been Model Predictive Control (MPC), which forward simulate the dynamics to predict the future states \citep{paden2016survey}. However, for this approach to work, a highly accurate and computationally efficient dynamics model is necessary. Furthermore, robotic systems deployed in the real world exhibit dynamical changes over time, and as such, these models must also adapt \citep{siciliano2016springer-changes}.

Traditionally for such systems, \textit{parametric models} are utilized based on known physics to which parameters are fitted from data. These types of models are explainable, efficient, and generalize well to large state spaces but usually do not capture all dynamical properties \citep{siciliano2016springer-param-models}. On the other hand, \textit{non-parametric models} that can use any function approximation technique to express the dynamics of the system. These data-driven models have the capability of capturing dynamics much more accurately and without requirements for extensive dynamics knowledge beforehand. However, these types of models are also not intuitively explainable and generalize poorly to unknown state spaces. A particular variety of non-parametric models that have seen success and popularity recently has been neural networks with several layers of depth, allowing them to capture highly non-linear dynamics. Finally, one of the more recent proposed categories of dynamics models has been semi-parametric models, which are a combination of the two models explained above \citep{nguyen2010using,romeres2016online,smith2020online}.

The model type and structure are important, but equally as crucial is learning their parameters. Traditionally this is done by collecting large datasets of real-world interactions, processing the data, learning the parameters from it, and then deploying the robotics system with these parameters fixed. This approach has successfully been used for decades. However, it is inherently limited when deployed outside a laboratory environment where the robot dynamics are continually changing from both wear and tear of the robot itself and external factors such as winds and surface friction. To address this problem, the dynamics model must be adapted online, which is a non-trivial problem due to the non-stationary learning target where any updates to the parametric part of the model modify the learning target of the non-parametric part \citep{smith2020online}. Furthermore, online learning is an even bigger issue when applied to neural networks due to the problem of catastrophic forgetting, which is their tendency to forget the old data when learning new ones \citep{robins1993catastrophic,goodfellow2013empirical,kirkpatrick2017overcoming}.

The novel contribution of this paper is applying an iterative learning non-linear semi-parametric model, with a neural network, to the task of autonomous racing with a Model Predictive  Controller (MPC).  We propose a process for bootstrapping the model from previously recorded data, deploying it, and then iteratively updating the model online. We experimentally show that the model successfully adapts to long-term changes in dynamics, achieves better performance than a purely parametric model, and generalizes better than a purely non-parametric model. The main contributions of this paper are:
\begin{enumerate}
    \setlength\itemsep{0em}
    \item A novel efficient semi-parametric model that utilizes a neural network that can be used as part of an MPC.
    \item An iterative learning system that adapts the model to modified dynamics while taking into account the previously encountered state space.
\end{enumerate}
\section{Related Work}
\label{sec:related-work}
Semi-parametric models are one of the more recently proposed models for dynamics learning and are a combination of parametric and non-parametric models. These models vary in form, but the general idea is that the parametric part captures the known (usually linear) structure of the dynamics, and the non-parametric part of the model captures the unmodeled dynamics (usually highly non-linear). This approach to modeling dynamics combines the generalization and efficiency capabilities of parametric models with the accuracy of non-parametric models. It has been successfully applied to various robotics platforms using different non-parametric functions, including Least Squares \citep{romeres2016online}, Gaussian Processes Regression (GPR) \citep{nguyen2010using} and Gaussian Mixture Models (GMM) \citep{smith2020online}.

Iterative learning (also known as online learning) for robotic systems has also seen an increase in interest recently due to its high desirability for real system deployment \citep{romeres2016online,smith2020online,williams2019locally}. Model adaption of purely parametric models is straightforward \citep{johansson1987parametric}, but doing so for semi-parametric models imposes the challenge of non-stationary target distributions. None the less, there exist examples of successful applications of this as a standalone model \citep{romeres2016online} and to Composite Adaptive Control \citep{smith2020online}. However, no approaches are using an iterative learning semi-parametric with a neural network applied to an MPC. The most notable similar work on this is applying a purely non-parametric neural network model to an iterative learning system using an MPC \citep{williams2019locally}, which serves as the basis to this work.

Iterative learning for neural networks has also received significant attention in recent years due to the rise in the popularity of deep learning. Many directions have been explored here, including controlling how parameters are updated \citep{kirkpatrick2017overcoming,wan2006parameter} and meta-learning, where there is another machine learning algorithm, is responsible for training the main model \citep{finn2017model,vilalta2002perspective}. However, the more mature family of techniques that explicitly tackle the issue of catastrophic forgetting is that of rehearsal method. The idea is that old data is stored and used to train the model in addition to the new data \citep{robins1993catastrophic}. However, maintaining all old data is infeasible, which has given rise to pseudo-rehearsal methods that do not explicitly store data but instead model it as data distribution and can generate random samples from it \citep{robins1995catastrophic}.

\section{Problem Formulation}
\label{sec:problem}

A robot operating in an environment with a model-based controller utilizes a dynamics model. The state $\textbf{x}_t$ and executed controls $\textbf{u}_t$ at the current time influence the state derivative $\dot{\textbf{x}}_t$. Assuming the model operates in discrete time with a timestep $\Delta t$, then then the state transition function is given by:
\begin{equation} \label{eq:dynamics_update}
    \textbf{x}_{t+1} =
    \textbf{x}_t + \dot{\textbf{x}_t} \Delta t = 
    \textbf{x}_t + \textbf{F}(\textbf{x}_t, \textbf{u}_t; \theta) \Delta t
\end{equation}
Where $\textbf{F}$ is the full system dynamics model with some parameters $\theta$. To learn the dynamics, our model must predict the state derivative $\dot{\textbf{x}}$. To obtain the discrete version of that and thus our prediction target $Y$ we reformulate Equation \ref{eq:dynamics_update} into:
\begin{equation} \label{eq:dynamics_target}
    Y_t =  \dot{\textbf{x}_t}
    = \textbf{F}(X; \theta)
    = \textbf{F}(\textbf{x}_t, \textbf{u}_t; \theta)
    = \frac{\textbf{x}_{t+1} - \textbf{x}_t}{\Delta t}
\end{equation}
where $X = (\textbf{x}_t; \textbf{u}_t)$ are the inputs into the model. For semi-parametric models, $\textbf{F}$ can be expanded into specific parametric and non-parametric parts. Without making any assumptions about the inputs to each, we rewrite Equation \ref{eq:dynamics_update} for a semi-parametric model:
\begin{equation} \label{eq:dynamics_update_semi}
    \textbf{x}_{t+1} = \textbf{x}_t + \left(\underbrace{f(X^f_t; \theta^f)}_{\text{parametric part}}
    + \underbrace{g(X^g_t; \theta^g)}_{\text{non-parametric part}}\right) \Delta t
\end{equation}
Note that in this general notation, both have different inputs and parameters. Assuming that the $X^g_t$ is a function of $X^f_t$, then the issue of non-stationary learning targets arises. To update the full model iteratively, both parts of the model need to be updated independently. Since neural networks are updated via batches, one can first update the parametric model and then fully train the neural network to account for all changes. However, this is not feasible for a control system, which must adapt to such changes in real-time. The closest to that is to update the neural network with only a few mini-batches, but that is not guaranteed to account for the changes in the parametric model. Thus this results in the non-parametric model trying to learn a constantly changing target distribution, which is challenging \citep{smith2020online}.

When updating the neural network parameters we also have to tackle the issue of catastrophic forgetting where if only newly collected data $X \sim \mathcal{P}_L(\mathcal{X})$ from the \textit{Local Operating Distribution} $\mathcal{P}_L(\mathcal{X})$ is used for learning, will result in the neural network forgetting prior dynamics. To tackle this with rehearsals \citep{robins1993catastrophic}, we assume that  another \textit{System Identification Distribution} $\mathcal{P}_{ID}(\mathcal{X})$ is available which captures data from all vehicle operations similar to \citep{williams2019locally}.


\section{Model}
\label{sec:model}

In this section, we describe the employed dynamics model. We chose to utilize a sequential semi-parametric modeling approach where the parametric part of the model first makes a state derivative prediction $\hat{\textbf{x}} = f(\textbf{x}, \textbf{u})$ which is then fed as an input into the non-parametric part of the model $g(\cdot)$ resulting in the final semi-parametric model represented as: 
\begin{align} \label{eq:generic_dynamic_bycicle_model}
    \dot{\textbf{x}} =
    \underbrace{
    f(\textbf{x}, \textbf{u})
    }_{\text{parametric part}}
    +
    \underbrace{
    g(\hat{\textbf{x}}, \textbf{u})
    }_{\text{non-parametric part}}
\end{align}
For our model, we define the state $\textbf{x}$ consisting of longitudinal and lateral position, yaw angle, longitudinal velocity, lateral velocity, and yaw rate; the control inputs $\textbf{u}$ are defined as acceleration and steering commands:
\begin{align}
    \textbf{x} = \begin{pmatrix}
    x &
    y &
    \psi &
    \dot{x} &
    \dot{y} &
    \dot{\psi}
    \end{pmatrix}
    &&
    \textbf{u} = \begin{pmatrix}
    u_a &
    u_\delta
    \end{pmatrix}
\end{align}
\subsection{Parametric part}
\label{sec:parametric}
Due to the inherent symmetry of cars, it is common to simplify 4-wheeled motors to a 2-wheeled one where the front and rear set of tires are lumped into only one at the front and one at the real. This is commonly referred to as the \textit{bicycle model} \citep{althoff2017commonroad}. The model can be summarized with the following state transition:
\begin{align}
\label{eq:dynamic_bycicle_model}
    \dot{\textbf{x}} = \begin{pmatrix}
    \dot{x} \\
    \dot{y} \\
    \dot{\psi} \\
    \ddot{x} \\
    \ddot{y} \\
    \ddot{\psi} \\
    \end{pmatrix} =
    \begin{pmatrix}
    sin(\psi) \dot{x} - sin(\psi) \dot{y}\\
    sin(\psi) \dot{x} + cos(\psi) \dot{y}\\
    \dot{\psi}\\
    \dot{\psi} \dot{y} + u_a \\
    -\dot{\psi} \dot{x} + \dfrac{2}{m} ( F_{yf} \cos{u_{\delta}} + F_{yr} ) \\
    \dfrac{2}{I_z} ( l_f F_{yf} - l_r F_{yr} )
    \end{pmatrix}
\end{align}
where $F_{yf}$ and $F_{yr}$ are the lateral tire forces at the front and rear respectively, and $m$, $I_{zz}$, $l_f$, $l_r$ are the mass, yaw inertia, distance from Center of Gravity (C.G.) to front axle and distance from C.G. to rear axle parameters respectively.

Except for aerodynamic drag and gravity, tire forces are the only external force acting on the vehicle. The tire forces have highly non-linear behavior at the limits of friction; however, it is also on that limit that tires exhibit maximum force, which is required for aggressive driving. Thus, correct tire modeling is essential for which we use the \textit{tire brush model}. In particular, the variant proposed by Fromm \citep{hadekel1952mechanical} which defines forces as:
\begin{align} \label{eq:tyre_brush_model}
    F_y = \begin{cases}
            -C_s \tan{\alpha} + \dfrac{C_s^2}{3 \mu F_z} | \tan{\alpha} | \tan{\alpha} - \dfrac{C_s^3}{27 \mu^2 F_{z}^2} \tan^3 \alpha  & |\alpha| \leq \alpha_{crit} \\
            -\mu F_{z} sgn (\alpha)                         & |\alpha| > \alpha_{crit}
    \end{cases}
\end{align}
where $C_{s}$ is the cornering stiffness, $\mu$ is the friction coefficient between the road surface and the tire, $\alpha$ is the slip angle of the front, and rear tires is given by:
\begin{align} \label{eq:slip_angle_alpha}
    \alpha_f = u_{\delta} - \arctan{\dfrac{\dot{y} + \dot{\psi}l_f}{\dot{x}}}
    &&
    \alpha_r = - \arctan{\dfrac{\dot{y} - \dot{\psi}l_r}{\dot{x}}}
    &&
    \alpha_{crit} = \arctan \dfrac{3 \mu F_z}{C_s}
\end{align}
$\alpha_{crit}$ is the smallest magnitude slip angle at which the entire contact patch is at the friction limit. $F_z$ is the vertical force for the tire which for our simplified model can be calculated for the front and rear axles with:
\begin{align} \label{eq:vertical_load}
    F_{zf} = \dfrac{l_r}{2(l_f + l_r)} m g
    &&
    F_{zr} = \dfrac{l_f}{2(l_f + l_r)} m g
\end{align}
where $m$ is the mass of the vehicle, and $g$ is the gravitational force.  Combining this tire model expressed by its lateral forces $F_{yf}$ and $F_{yr}$ with Equation \ref{eq:dynamic_bycicle_model} results in our final parametric model.

\subsection{Non-parametric part}
\label{sec:non-parametric}
\begin{wrapfigure}{R}{0.5\textwidth}
    \centering
    \includegraphics[width=\linewidth]{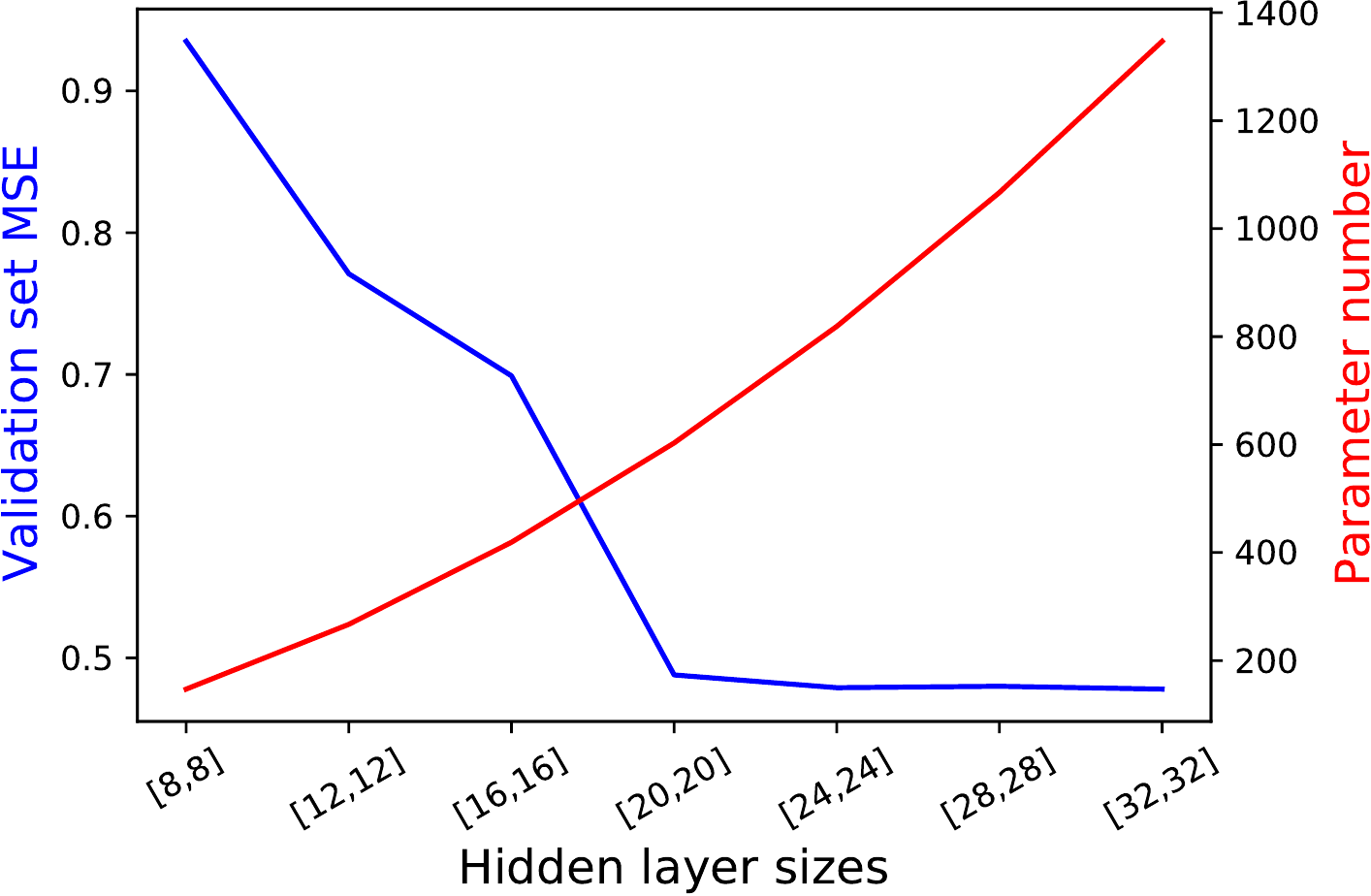}
    \caption{Accuracy comparison between networks of different hidden layer sizes.}
    \label{fig:network-sizes}
\end{wrapfigure}
Intending to enhance the performance of the parametric model, we chose to use a neural network with 2 hidden layers as it can accurately capture the highly non-linear dynamics missed by the parametric model and do so with excellent computational efficiency, as shown in \citep{williams2019locally}. For the purposes of this project, the non-parametric part predicts only the velocities and yaw rate. In Figure \ref{fig:network-sizes}, it was experimentally found that a network with hidden layer sizes of (20, 20) and $tanh$ activation functions performs sufficiently well while maintaining a small set of parameters. For the other hyper-parameter, we also experimentally identified that the following parameters produce the best results: batch size of 100, epoch size of 1000, Adam optimizer \citep{kingma2014adam} with a learning rate of $10^{-3}$, L2 regularization with a weight decay coefficient of $10^{-3}$ and data normalization. The reasoning for these choices can be found in Appendix \ref{app:semi-param-hyper-search}. We found it particularly useful to use significant L2 regularization when training the full semi-parametric model. The reasoning for this is that the neural network is only learning residuals and does not need to make significant corrections to the prior prediction of the parametric part of the model.
\section{Iterative learning}
\label{sec:iterative-learning}

In designing this system, we had two goals in mind (1) the model must adapt to changes swiftly but also (2) not wholly ignore prior dynamics. The justification for this is that the system must adapt to changes quickly but also be resilient to short-term disturbances such as wind gusts. We first assume that the model is bootstrapped with data generated from the $\mathcal{P}_{ID}(\mathcal{X})$ distribution, where the parametric part of the model is fit first, and the neural network learns its residuals. Then we assume that the dynamics remain similar throughout the operation of the robot and that the parametric model parameters will not change significantly; as such, we chose only iteratively to update the neural network. Although this is a strong assumption, it guarantees that we do not encounter the problem of non-stationary targets (Section \ref{sec:problem}) while still learning an accurate dynamics model, albeit potentially slower.

\begin{figure}[H]
    \centering
    \includegraphics[width=\linewidth]{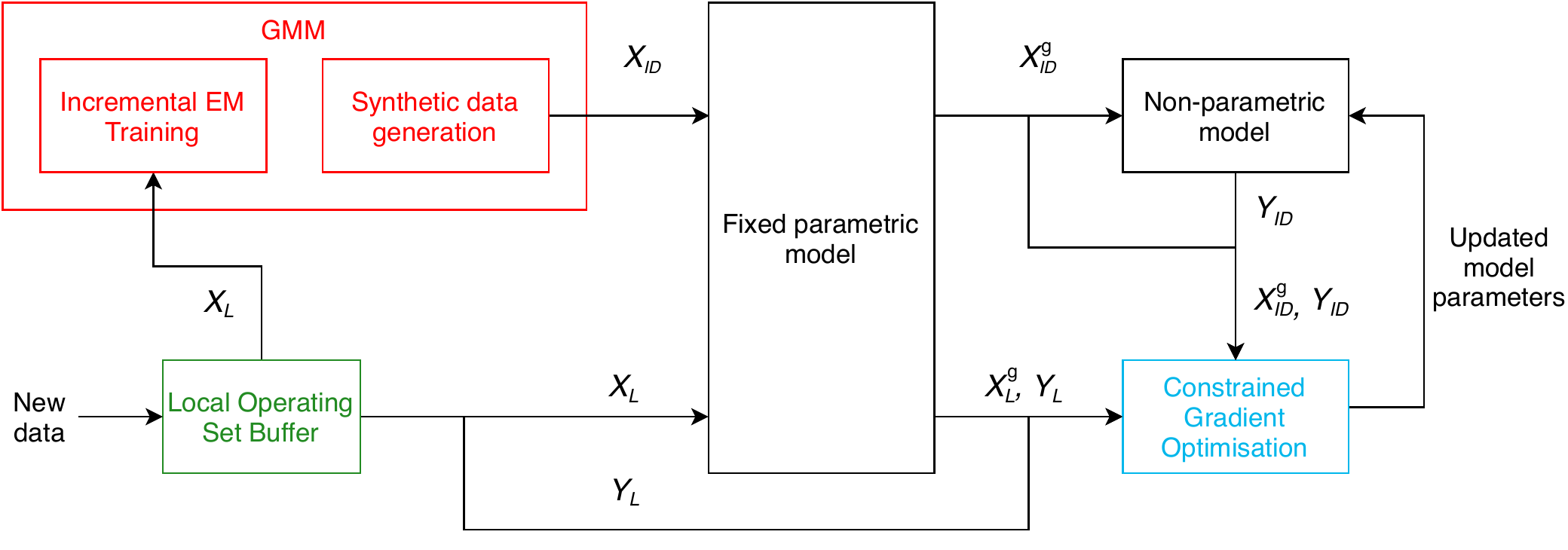}
    \caption{Iterative learning system for the non-parametric model.}
    \label{fig:iterative_learning}
\end{figure}

To address the issue of catastrophic forgetting in the neural network (Section \ref{sec:problem}), we chose to use pseudo-rehearsal methods where the model is trained simultaneously on the old data (drawn from $\mathcal{P}_{ID}(\mathcal{X})$) and the new data (drawn from $\mathcal{P}_{L}(\mathcal{X})$). For generating inputs $X_{ID} \sim \mathcal{P}_{ID}(\mathcal{X})$, a {\color{Red}Gaussian Mixture Model (GMM)} \citep{reynolds2009gaussian} is used, and the neural network generates the corresponding targets $Y_{ID}$. The GMM is fit using expectation-minimization (EM) \citep{dempster1977maximum}. To guarantee consistency between the two data distributions, we utilize a {\color{Cyan} constrained gradient update rule} introduced in \citep{williams2019locally}. The similarity between the two data distributions is enforced via the cosine of the angle between the update directions of the gradients computed from both datasets:
\begin{align}
    \theta_{i+1}^g &= \theta_i^g - \gamma(\alpha G_L + G_{ID})\\
    \alpha &= \underset{a \in [0,1]}{\max} \text{      s.t. } \langle a G_L + G_{ID}, G_{ID} \rangle \ge 0
\end{align}
where $\gamma$ is the learning rate, $G_L$ and $G_{ID}$ are the gradients of the model parameters computed from the two data distributions:
\begin{align}
    G_L &= \nabla_{\theta_i^g} \lVert Y_L - g(X_L^g, \theta_i^g)\rVert^2\\
    G_{ID} &= \nabla_{\theta_i^g} \lVert Y_{ID} - g(X_{ID}^g, \theta_i^g)\rVert^2
\end{align}
The new data is stored in the {\color{ForestGreen} Local Operating Set Buffer} of fixed length and is used as a full batch along with a batch of equal size generated by the GMM to train the neural network. Training initiates online when the {\color{ForestGreen} buffer} is full of new data. We empirically found that it is best to use the same optimizer and hyper-parameters as used during the bootstrapping. After training has finished, we use the new data to {\color{Red} update the GMM via an incremental EM training approach} \citep{zhang2010online}. This way, the memory footprint of the data is kept low while ensuring the model never forgets old data. A visual guide of the system is presented in Figure \ref{fig:iterative_learning}.
\section{Experimental Validation}
\label{sec:result}
\subsection{Experimental Setup}

We test our approach to the task of learning dynamics as part of an MPC in an autonomous racing scenario where the goal is to drive as fast as possible without crashing or leaving the race track boundaries. This is a particularly difficult problem as aggressive driving attempts to exploit the dynamical limits of the vehicle, which are highly non-linear, rarely observed, and challenging to model.

The target platform for our experiments is a full-scale electric vehicle fitted with all the necessary sensors and compute for the task. Dynamically the vehicle is capable of achieving 11,8 m/s\textsuperscript{2} lateral acceleration, 4,9 m/s\textsuperscript{2} longitudinal acceleration, 8,8 m/s\textsuperscript{2} longitudinal deceleration (braking) and top speeds of 200 kph (56 m/s). Sadly due to external factors, we were unable to execute real-world experiments, and as such, this section is only limited to simulated ones. A proprietary simulator is used with an extremely accurate vehicle dynamics model. Bootstrapping and validation data is collected from this simulation by manual driving to maintain real-world feasibility. The data is then filtered, smoothed, and prepared for model learning.

For the remainder of this section, we have conducted three types of experiments. (1) \textit{semi-parametric model experiments} which aim to validate the benefits of semi-parametric models. (2) \textit{offline iterative learning experiments} where the system is bootstrapped on a dataset and then trained iteratively on another dataset collected with modified dynamics. We evaluate the initial and adapted models on the latter dataset. (3) \textit{online iterative learning experiments} where the system is bootstrapped on a dataset and then deployed along with a controller in the simulation with modified dynamics. For this, we use the open-source Model Predictive Path Integral (MPPI) controller, which is a sampling-based, gradient-free method that computes the optimal trajectory as a cost-weighted average over randomly sampled trajectories \citep{williams2018information}.

\subsection{Semi-parametric Model Experiments}
\label{sec:semi-parametric-experiments}
\begin{wrapfigure}{R}{0.5\textwidth}
    \centering
    \vspace{-1.1cm}
    \includegraphics[width=\linewidth]{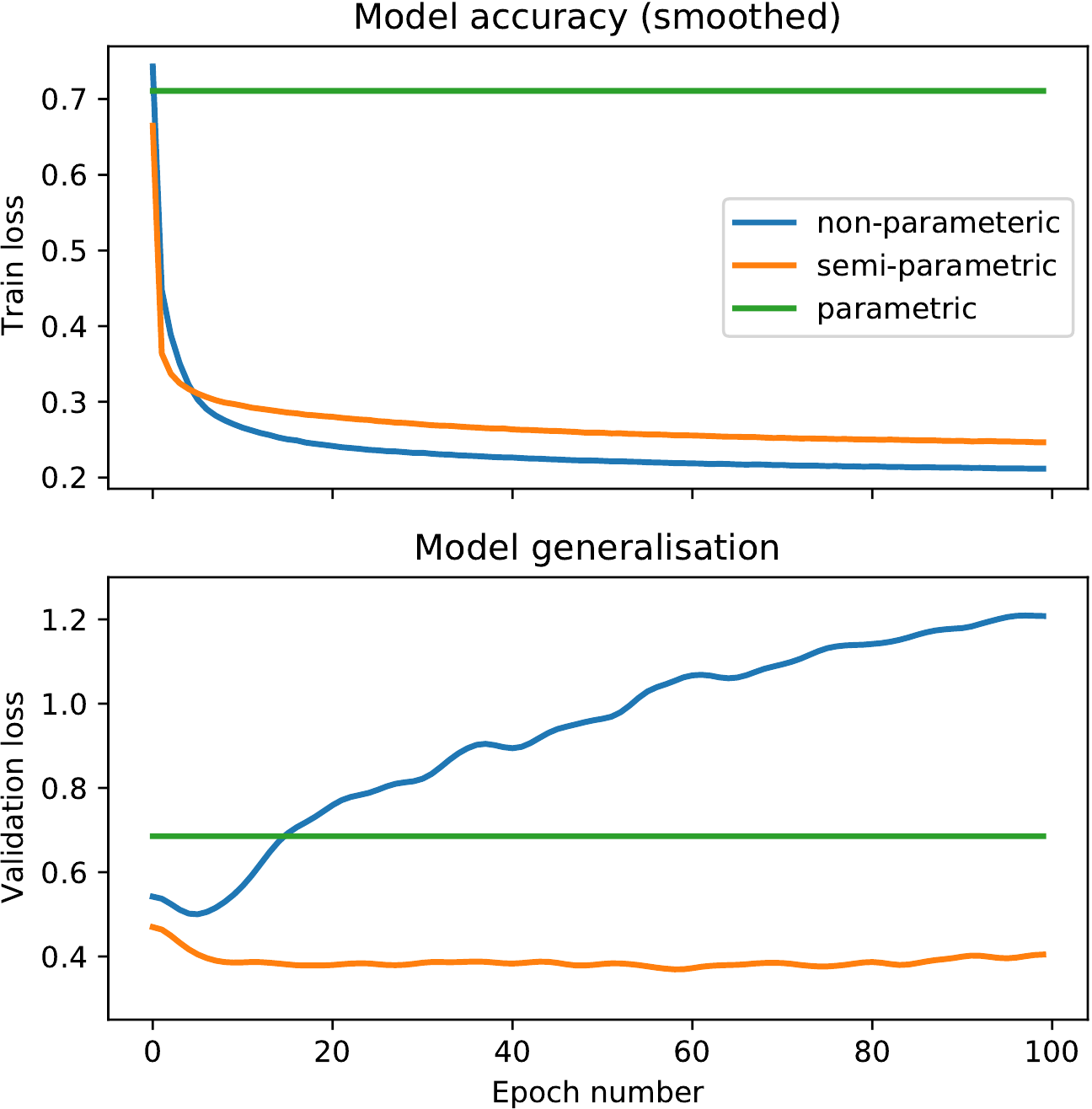}
    \caption{Comparison between the learning and generalization capabilities of the different model types The purely parametric model shows a constant error since it is not trained in batches.}
    \label{fig:model_types}
\end{wrapfigure}
In this subsection, we validate the generalization and efficiency of using semi-parametric models. Firstly, we split the dataset collected into a train, validation, and test sets such that each set contains data from a unique part of the state space. This is done by sorting all the data by longitudinal velocity data and splitting it into the sets as proportions of the full spectrum of velocities. This is a plausible real-world scenario as it is usually not feasible and expensive to collect large datasets exploring the entire system state space. Further explanation and visualizations can be found in Appendix \ref{app:data-split}. Next, we evaluate the generalization capabilities of the semi-parametric model by comparing it to only its parametric part and only its non-parametric parts. The results of this training procedure plotted against the neural network epoch number can be seen in Figure \ref{fig:model_types}.

The purely parametric model exhibits the worst training set loss. However, taking a look at the validation loss performance, the purely parametric model shows superior generalization capabilities when compared to the strictly non-parametric model, which overfits the training state space. The semi-parametric model achieves similar accuracy to the purely non-parametric model on the training set and is slightly more efficient at learning while also showing excellent generalization capabilities on the validation set. It is worth noting that here for the purely non-parametric model, we use a larger neural network with 2 hidden layers of 32 neurons each as otherwise, the model is not able to capture the dynamics adequately.

\subsection{Offline Iterative Learning Experiments}
\label{sec:offline-experiments}
In this subsection, we use the bootstrapping and validation datasets described earlier to evaluate the proposed iterative learning offline without the added stochasticity of introducing a controller. We first use the bootstrapping dataset to train the model and then let it adapt to the changed dynamics in the validation set where the mass of the vehicle changed from 1350 kg to 1430 kg and the friction coefficient $\mu$ from 1,0 to 0,8. This is a plausible real-world scenario found during empirical testing where a passenger can increase the overall mass of the vehicle, and a different road surface can reduce tire friction. After the model has adapted through the whole dataset, we evaluate its prediction accuracy on the bootstrapping data, on the validation data, and finally on a third dataset, which we will call test data consisting again of modified dynamics. Empirically we found that a local operating set of 500 data entries (10 seconds of driving) worked well combined with a mini-batch size of 100 and 3 epochs of training every time the local operating set is full. To highlight the generalization capabilities of our proposed approach, we also show the results of using pure Stochastic Gradient Decent (SGD) for learning with the same hyper-parameters as our proposed approach. The final results are shown in Table \ref{tab:offline-experiments}, where we can see that both of the iterative learning models adapt successfully to the modified dynamics of the validation dataset; however, the performance of the simple SGD approach deteriorates when evaluated on the test data. On the other hand, our proposed iterative learning approach produces excellent results on the validation data while also generalizing to the test data.
\begin{table}[h]
\vspace{-0.25cm}
\centering
\caption{Offline iterative learning experimental results.}
\label{tab:offline-experiments}
\begin{tabular}{@{}lp{2cm}p{2cm}p{2cm}p{2cm}@{}}
\toprule
                     & Bootstrap data MSE & Validation data MSE & Test data MSE \\ \midrule
Bootstrapped         & 0.504          & 1.037           & 1.050  \\
SGD                  & 2.148          & 0.121           & 0.442  \\
Iterative learning   & 1.423          & 0.181           & 0.170  \\
\end{tabular}
\end{table}
\vspace{-0.25cm}
\subsection{Online Iterative Learning Experiments}
\label{sec:online-experiments}
In this subsection, we deploy the semi-parametric model and the iterative learning system as part of the MPPI controller. First, we use the trained semi-parametric model from Section \ref{sec:semi-parametric-experiments} without iterative learning; we call this the \textit{normal} scenario. Then we run another experiment with a simulation with the same dynamics modifications introduced in the previous subsection; we call this the \textit{modified} experiment. We chose the control parameters to be safe ones to ensure that all models complete the course successfully. These two experiments serve as the lower and upper bound of performance we should expect. Finally, we run our iterative learning system bootstrapped with data of unmodified dynamics. Figure \ref{fig:iterative_learning_experiments} shows the results of 10 consecutive laps of driving averaged over five different experiments for each scenario to account for the stochastic nature of the controller and the dynamics learning system. Table \ref{tab:it_errors} shows one of the iterative learning experiments in more detail where the MSE is calculated on the dataset of modified dynamics collected over 10 laps. To showcase the model learning, we have saved snapshots of the model at the beginning of an experiment (0 s), at 60 s, and 120 s of learning and ran these as fixed models for one lap. Then we plot the resulting GG diagrams in Figure \ref{fig:gg-plot}, which indicate how well the controller harnesses the full vehicle dynamics.

\begin{multicols}{2}
\begin{Figure}
    \centering
    \includegraphics[width=\linewidth]{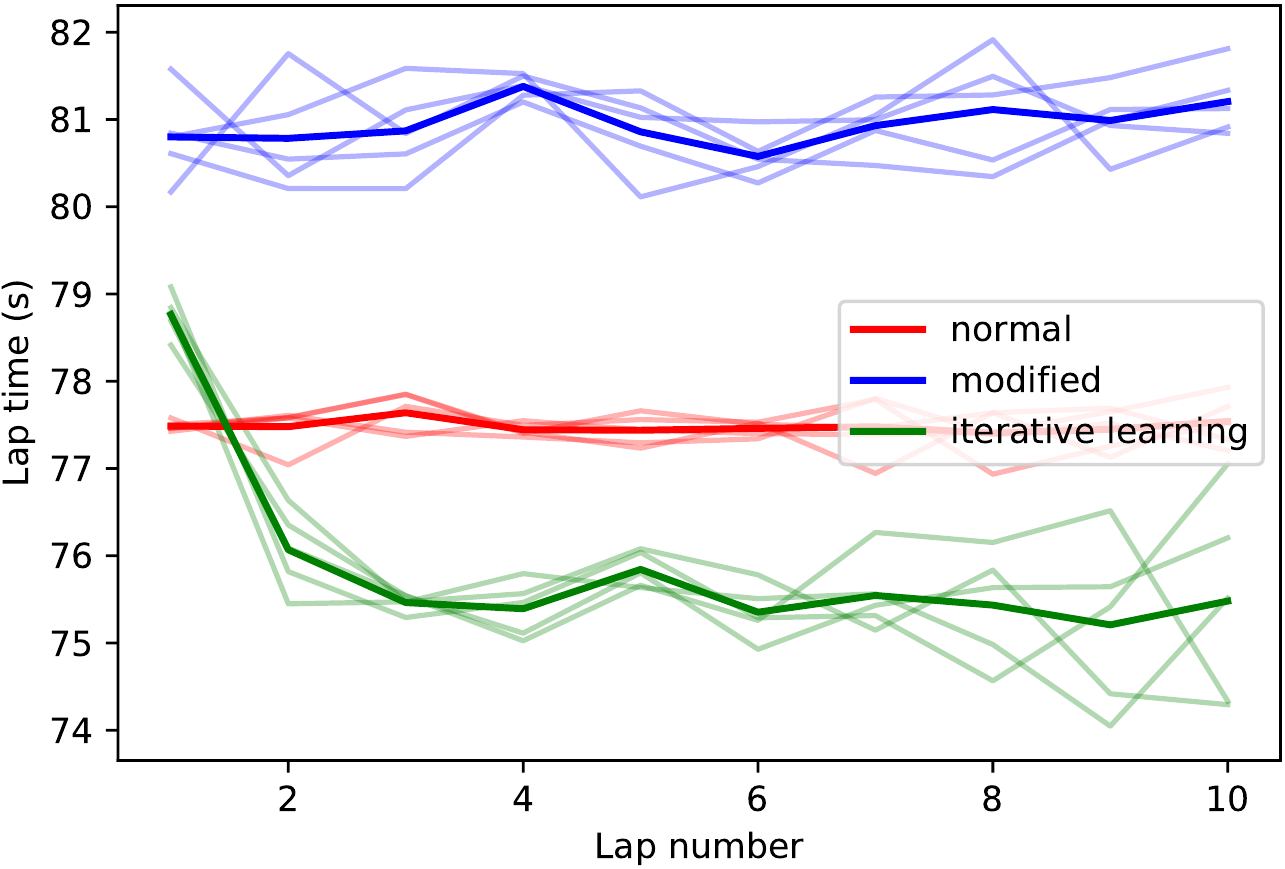}
    \captionof{figure}{Consecutive lap times of iterative learning system compared to the fixed models. Each model is ran for 5 different experiments, and their average is shown with a solid line.}
    \label{fig:iterative_learning_experiments}
\end{Figure}
\begin{table}[H]
    \centering
    \begin{tabular}{@{}lll@{}}
    \toprule
    Lap & Lap Time & Model MSE    \\ \midrule
    0   & -      & 1.050    \\
    1   & 78.69  & 0.210    \\
    2   & 75.93  & 0.194    \\
    3   & 75.52  & 0.193    \\
    4   & 75.53  & 0.192    \\
    5   & 74.94  & 0.194    \\
    6   & 75.55  & 0.192    \\
    7   & 75.55  & 0.192    \\
    8   & 75.44  & 0.192    \\
    9   & 75.03  & 0.193    \\
    10  & 75.24  & 0.193    \\
    \end{tabular}
    \caption{Lap time performance of the iterative learning system and model MSE over one of the 10-lap experiments from the Figure on the left. Lap 0 stands for the bootstrapped model.}
    \label{tab:it_errors}
\end{table}
\end{multicols}

From these results, we can see that iterative learning is working successfully and adapts to the changing dynamics\footnote[1]{A supplementary video is available online - \url{https://youtu.be/aLWXa95AEKE}.}. Furthermore, it consistently outperforms the \textit{normal} experiments, which were previously considered the upper bound of performance. This could be because the model fits more specifically to the current racetrack or because the controlled explores the state space better than the human driver.  However, there is also a negative effect - the iterative learning model becomes progressively more unstable, which is shown via the variance in the lap times of the iterative learning system. We do not explicitly consider comparisons to simple SGD model updates as those usually result in incorrectly learned dynamics and the controller being unable to finish a single lap.

\begin{multicols}{2}
\begin{Figure}
    \centering
    \includegraphics[width=0.9\linewidth]{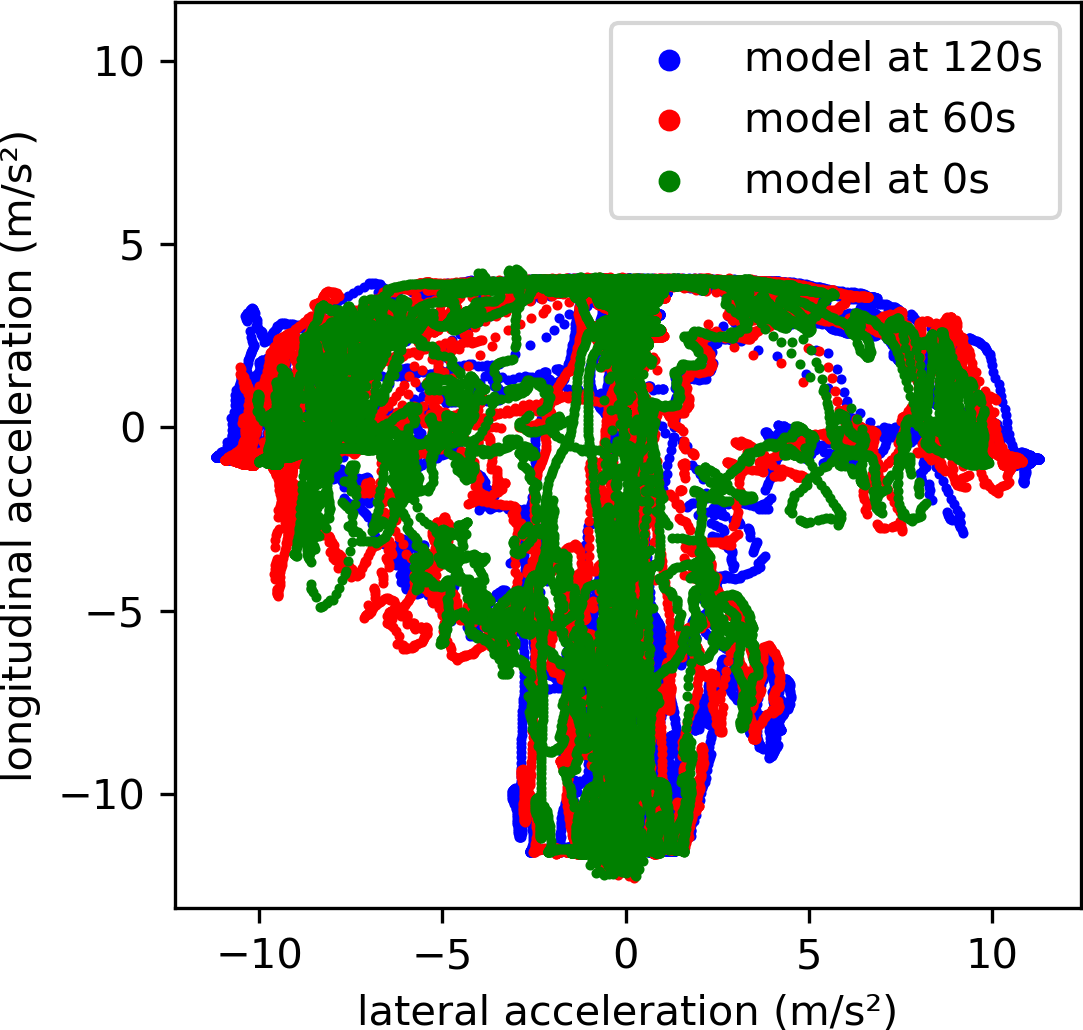}
    \captionof{figure}{GG plot of accelerations achieved by the iteratively learned model at different stages. Larger lateral acceleration magnitudes indicate better and more aggressive performance.}
    \label{fig:gg-plot}
\end{Figure}

\begin{Figure}
    \centering
    \includegraphics[width=0.85\linewidth]{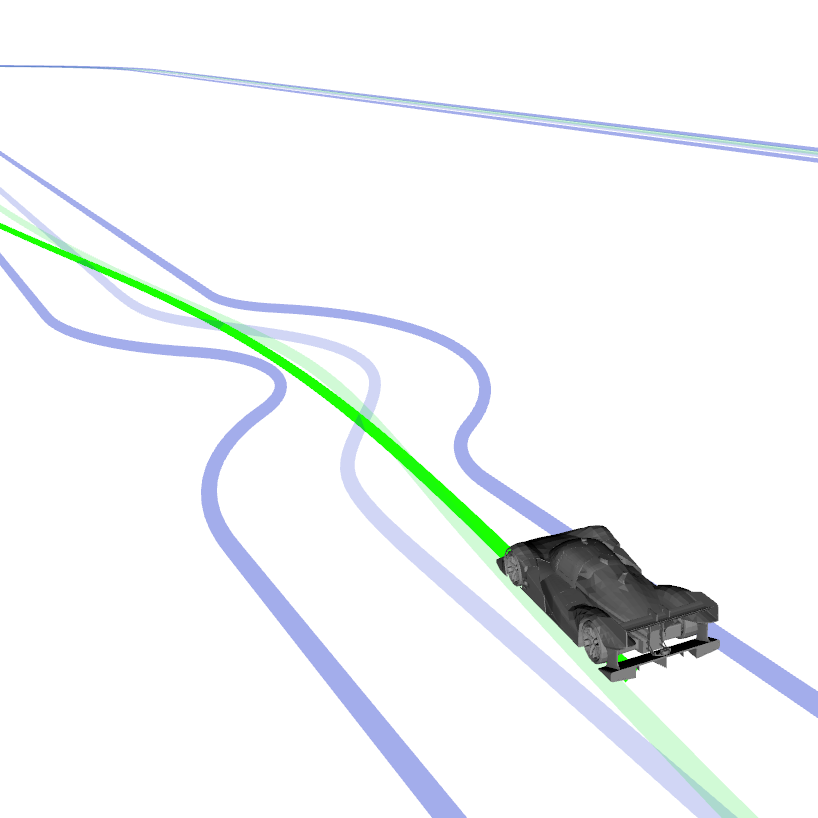}
    \captionof{figure}{Data visualization from the simulation. The solid green line is the optimized trajectory by MPPI, and the transparent green line is the reference path.}
    \label{fig:simulation}
\end{Figure}
\end{multicols}

\section{Conclusion}
\label{sec:conclusion}

In this paper, we present a novel approach to modeling robot dynamics by using a non-linear semi-parametric model with a neural network, resulting in an accurate and computationally efficient model. We developed a system for iteratively updating the non-parametric part of the model using pseudo-rehearsals and apply it to the case of Model Predictive Control (MPC) in autonomous racing. We show that our model achieves excellent accuracy in comparison to purely parametric models while also retaining superb generalization capabilities in contrast to a strictly non-parametric model. To test our iterative learning system, we use several datasets to evaluate our approach and show that our model overcomes the issue of catastrophic forgetting common for neural networks. Furthermore, it adapts to changing dynamics with batch learning and also generalizes to unseen state spaces. Finally, we deploy our system along with an MPC in a simulation and show that our system can safely adapt changing dynamics at operation time and improve overall controller performance.

Due to external circumstances, we were unable to apply our approach to a real system. However, we would like to note that our iterative learning system allows for straightforward simulation-to-real-world transfer where the model can be bootstrapped in simulation and then left to adapt in the real-world iteratively. Even though our simulation shows positive results, dynamics modeling with neural network has two short-comings - instability and batch-learning. The first issue is caused by data overfitting and exploding gradients, which, although can be mitigated with regularization techniques, is still an underlying concern when deploying systems with long life-spans. The second issue of batch-learning is inherent to neural networks. It renders them incapable of learning short-term changes to the system dynamics, which is desirable but of uncertain feasibility. Additionally, in our approach, we do not update the parametric part of the model iteratively and leave its error to be absorbed into the non-parametric part of the model. Updating the full semi-parametric model is desirable but difficult \citep{smith2020online}.


\clearpage
\acknowledgments{The MPC used throughout this paper is based on the open-source implementation of the Model Predictive Path Integral (MPPI) controller by G. Williams et al. \citep{williams2016aggressive}}

\bibliography{example.bib}  

\clearpage
\begin{appendices}

\section{Semi-parametric model hyper-parameter search}
\label{app:semi-param-hyper-search}
Due to that and computational restrictions, we chose to explore hidden layer sizes of (8,8), (12,12), (16,16), (20,20), (24,24), (28,28) and (32,32) and learning rates of 0.0025, 0.001 and 0.00075. The other hyper-parameters are shown below, and the results are shown in Table \ref{tab:semi-paramteric-experiments}. 
\begin{itemize}
    \setlength\itemsep{0em}
    \item batch size of 100 and epoch size of 1000
    \item tanh activation function
    \item Adam optimizer 
    \item $10^{-5}$ weight decay coefficient
    \item data normalization only for the non-parametric part
    \item 70\%, 20\%, 10\% train, validation and test dataset split
\end{itemize}

\begin{table}[H]
\centering
\caption{Semi-parametric training experiments with different hidden layer sizes and learning rates. Average lap times are estimated over five consecutive laps. A failure is designated if the car span or went off track; in that case, lap times are omitted from the table. The best hyper-parameters for each model size is shown in bold. The model which has achieved the best lap times is italicized.}
\begin{tabular}{@{}p{1.4cm}p{1.5cm}p{1.4cm}p{1.2cm}p{1.2cm}@{}}
\toprule
Hidden layer size       & Learning rate           & Validation MSE           & Average lap time           & Lap time std.            \\ \midrule
8,8                   & 0.0025                  & 0.981                  & -                & - \\
\textbf{8,8}          & \textbf{0.001}          & \textbf{0.935}         & -                & - \\
8,8                   & 0.00075                 & 1.044                  & -                & - \\
12,12                 & 0.0025                  & 0.801                  & -                & - \\
\textbf{12,12}                 & \textbf{0.001}          & \textbf{0.771}         & \textbf{126.45}  & \textbf{0.744} \\
12,12                 & 0.00075                 & 0.794                  & -                & - \\
16,16                 & 0.0025                  & 0.733                  & -                & -    \\
\textbf{16,16}        & \textbf{0.001}          & \textbf{0.699}         & \textbf{124.82}  & \textbf{0.24} \\
16,16                 & 0.00075                 & 0.7368                 & 125.06           & 0.13 \\
20,20                 & 0.0025                  & 0.502                  & 126.2583         & 0.8365 \\
\textbf{20,20}        & \textbf{0.001}          & \textbf{0.491}         & \textbf{120.65}  &\textbf{0.35} \\
20,20                 & 0.00075                 & 0.496                 & 121.98            & 0.484 \\
24,24                 & 0.0025                  & 0.508                 & 126.98            & 0.836 \\
\textbf{24,24}        & \textbf{0.001}          & \textbf{0.479}        & \textbf{121.12}   & \textbf{0.348} \\
24,24                 & 0.00075                 & 0.491                 & 121.27            & 0.461 \\
28,28                 & 0.0025                  & 0.511                 & -                 & - \\
\textbf{28,28}                 & \textbf{0.001}                   & \textbf{0.480}        & \textbf{121.64}   & \textbf{0.488} \\
28,28                 & 0.00075                 & 0.485                 & 120.88            & 0.374 \\
32,32                 & 0.0025                  & 0.525                 & -                 & -    \\
32,32                 & 0.001                   & 0.479                 & 125.6762          & 0.6672 \\
\textbf{32,32}        & \textbf{0.00075}        & \textbf{0.478}        & \textbf{121.93}   & \textbf{0.248} \\
\bottomrule
\end{tabular}
\label{tab:semi-paramteric-experiments}
\end{table}

We can see that the bolded model with hidden layer sizes (20, 20) achieves sufficiently good performance in both Validation MSE and average lap times while maintaining computational efficiency. However, we are still not finished with identifying hyper-parameters. The semi-parametric model differs from the non-parametric model as in that it learns not the full model, but only the residuals leftover from the parametric part. As such, one would expect the weights of such a model to be smaller. Thus, here we consider one more important hyper-parameter - the \textit{weight decay coefficient}, which is responsible for the regularization of the neural network. In the case of this project, we use L2 regularization due to its prior availability and excellent results. We put this to the test with the previously identified best semi-parametric model and show the results in Table \ref{tab:non-parametric-wd}. From there, we can identify two good values of the parameter: $\lambda = 10^{-4}$ produces the best result, but $\lambda = 10^{-3}$ provides more stable lap times as training error plateaus. As stability is crucial for the vehicle model, we chose to utilize the latter.

\begin{table}[H]
    \centering
    \caption{Comparison between the model trained with different weight decay coefficients.  Average lap times are estimated over five consecutive laps. A failure is designated if the car span or went off track; in that case, lap times are omitted from the table.}
    \begin{tabular}{@{}p{1.9cm}p{1.4cm}p{1.2cm}p{1.2cm}@{}}
    \toprule
    Weight decay coeff. ($\lambda$)  & Validation MSE   & Average lap time & Lap time std. \\ \midrule
    $10^{-1}$             & 1.721            & -       & - \\
    $10^{-2}$             & 0.951            & 128.73  & 2.19 \\
    $10^{-3}$             & 0.504            & 121.12  & 0.12 \\
    $10^{-4}$             & 0.488            & 120.65  & 0.35 \\
    $10^{-5}$             & 0.491            & 120.94  & 0.47 \\
    
    \bottomrule
    \end{tabular}
    \label{tab:non-parametric-wd}
\end{table}

\section{Data split for different state spaces}
\label{app:data-split}

\begin{figure}[H]
    \centering
    \includegraphics[width=0.8\linewidth]{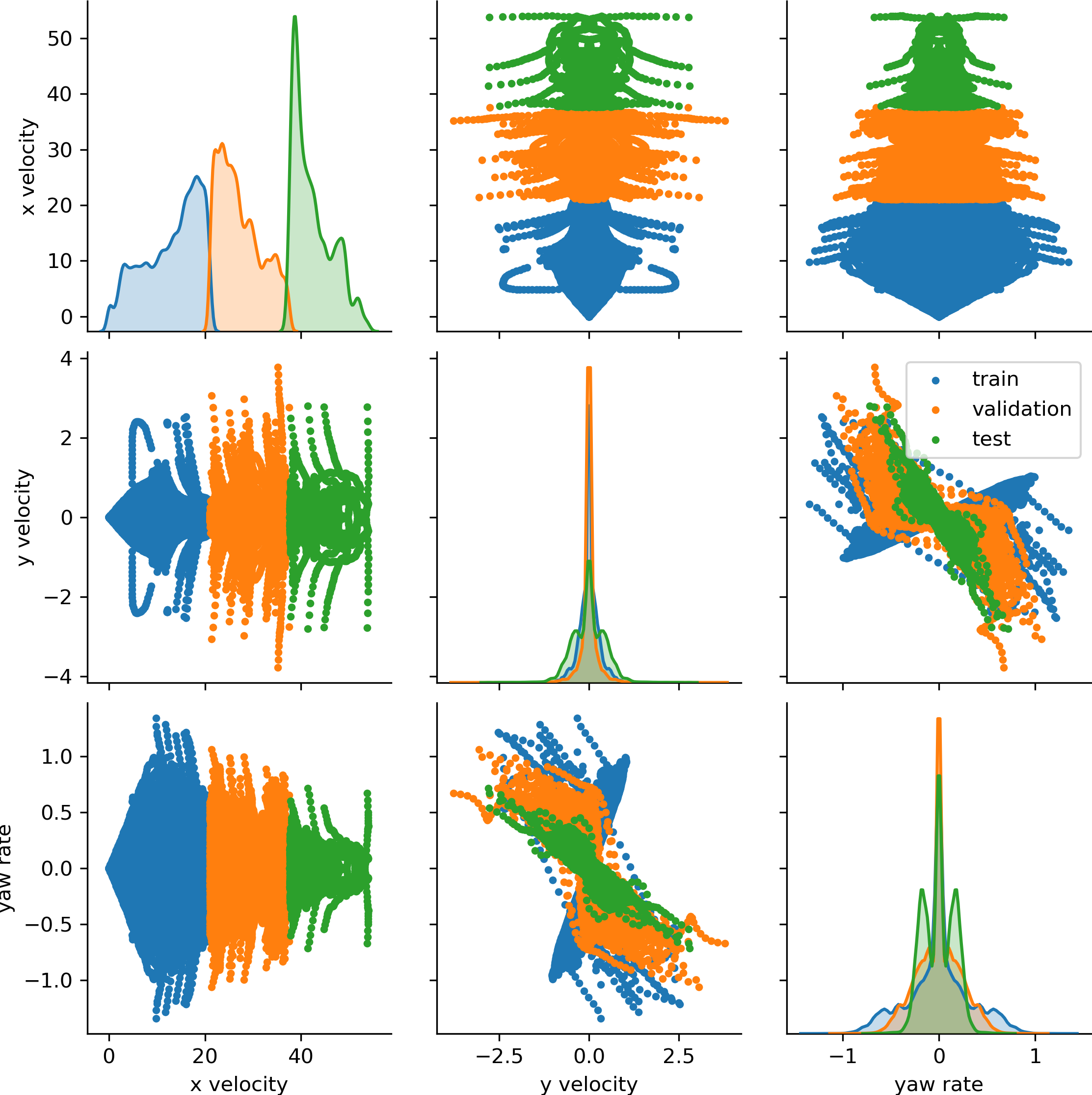}
    \caption{Example of how the dataset is split into train, validation, and test sets such that they occupy a different part of the overall state space. In this particular example, the data is divided according to the x velocity field, which can be seen in the top left plot. In this plot, the train, validation, and test datasets have 60\%, 35\%, 5\% proportions. It is worth noting that these proportions are not accurately represented in the diagonal kernel density plots.}
    \label{fig:real_validation}
\end{figure}

To test the generalization and accuracy benefits of the semi-parametric model in Section \ref{sec:semi-parametric-experiments}, we had to modify the typical machine learning validation process where data is randomized into train, validation and test sets. Instead, we split our data based on sorted longitudinal velocities, which nicely separates the whole state space such that the validation set includes data that is not present in the test set. Thus, this approach allows us to validate the generalization capabilities of dynamics models and is reflective of real-world applications where obtaining data from the full state space is not feasible. An example of this is shown in Figure \ref{fig:real_validation}.

\end{appendices}

\end{document}